\pdfoutput=1
\documentclass[runningheads]{llncs}

\usepackage[T1]{fontenc}
\usepackage{graphicx}
\usepackage{pdfpages}
\graphicspath{{figures/}}

\usepackage{amsmath,amssymb}
\usepackage{booktabs}
\usepackage{longtable}
\usepackage{xcolor}
\PassOptionsToPackage{hyphens}{url}
\usepackage{hyperref}
\usepackage{enumitem}
\usepackage{placeins}
\usepackage{silence}
\usepackage{subcaption}

\hypersetup{colorlinks=true, linkcolor=blue, citecolor=blue, urlcolor=blue}

  \newcommand{\selfcitepred}{our prior model~\cite{grimbly2026interoceptive}}

\newif\ifSUBMISSIONONLY
\newif\ifAPPENDIXONLY
\ifdefined\APPENDIXBUILD
  \APPENDIXONLYtrue
  \SUBMISSIONONLYfalse
\else
  \APPENDIXONLYfalse
  \ifdefined\FULLBUILD
    \SUBMISSIONONLYfalse
  \else
    \SUBMISSIONONLYtrue
  \fi
\fi

\newcommand{\baseuniform}{uniform baseline}

\newcommand{\kapparoute}{$\kappa$-routing}
\newcommand{\dynC}{dynamic-$C$}

\newif\ifCONFIRMATORYREADY
\IfFileExists{confirmatory_results.tex}{%
  \CONFIRMATORYREADYtrue
\newcommand{\ConfirmBranch}{C}

\newcommand{\ConfirmPrimaryMean}{-0.008}
\newcommand{\ConfirmPrimaryCILow}{-0.025}
\newcommand{\ConfirmPrimaryCIHigh}{+0.016}
\newcommand{\ConfirmPrimaryP}{0.5793}
\newcommand{\ConfirmModerationStatus}{estimated}

\newcommand{\ConfirmMarginalKappaMean}{+0.111}
\newcommand{\ConfirmMarginalKappaCILow}{+0.076}
\newcommand{\ConfirmMarginalKappaCIHigh}{+0.147}

\newcommand{\ConfirmMarginalKappaHolmP}{<0.0001}
\newcommand{\ConfirmMarginalCMean}{+0.080}
\newcommand{\ConfirmMarginalCCILow}{+0.046}
\newcommand{\ConfirmMarginalCCIHigh}{+0.118}

\newcommand{\ConfirmMarginalCHolmP}{<0.0001}
\newcommand{\ConfirmInteractionMean}{+0.114}
\newcommand{\ConfirmInteractionCILow}{+0.049}
\newcommand{\ConfirmInteractionCIHigh}{+0.182}

\newcommand{\ConfirmInteractionHolmP}{0.002}

}{\CONFIRMATORYREADYfalse}
\newif\ifCONFIRMBRANCHA
\newif\ifCONFIRMBRANCHB
\newif\ifCONFIRMBRANCHC
\newif\ifCONFIRMMODESTIMATED
\newif\ifCONFIRMMODNOINTERVAL
\newif\ifCONFIRMMODUNDEFINED
  \newcommand{\ConfirmBranchAToken}{A}
  \newcommand{\ConfirmBranchBToken}{B}
  \newcommand{\ConfirmBranchCToken}{C}
  \ifx\ConfirmBranch\ConfirmBranchAToken
    \CONFIRMBRANCHAtrue
  \else\ifx\ConfirmBranch\ConfirmBranchBToken
    \CONFIRMBRANCHBtrue
  \else\ifx\ConfirmBranch\ConfirmBranchCToken
    \CONFIRMBRANCHCtrue
  \else
    \PackageError{paper2}{Unknown confirmatory narrative branch}{Expected A, B, or C.}
  \fi\fi\fi
  \newcommand{\ConfirmModEstimatedToken}{estimated}
  \newcommand{\ConfirmModNoIntervalToken}{estimated-without-bootstrap-interval}
  \newcommand{\ConfirmModUndefinedToken}{undefined}
  \ifx\ConfirmModerationStatus\ConfirmModEstimatedToken
    \CONFIRMMODESTIMATEDtrue
  \else\ifx\ConfirmModerationStatus\ConfirmModNoIntervalToken
    \CONFIRMMODNOINTERVALtrue
  \else\ifx\ConfirmModerationStatus\ConfirmModUndefinedToken
    \CONFIRMMODUNDEFINEDtrue
  \else
    \PackageError{paper2}{Unknown confirmatory moderation status}{Expected an audited status token.}
  \fi\fi\fi

\newcommand{\authorcid}[1]{}

\newif\ifCODEAVAIL
\CODEAVAILtrue
\newcommand{\codeurl}{https://github.com/sgrimbly/interoceptive-prioritisation-iwai2026-snapshot}

\hypersetup{pdftitle={Two Routes to Interoceptive Prioritisation: Dissociating Likelihood Precision and Preference Reweighting in Active Inference}, pdfauthor={St John Grimbly, Nicolas Kuske, Evert A. Boonstra, Charel van Hoof, Rowan Hodson, Benjamin Rosman, Ryan Smith, Mark Solms, Jonathan P. Shock}}
\begin{document}

\title{Two Routes to Interoceptive Prioritisation: Dissociating
       Likelihood Precision and Preference Reweighting in Active Inference}
\titlerunning{Two Routes to Interoceptive Prioritisation}

\author{
St John Grimbly\inst{1,2}\authorcid{0000-0002-3003-1815}\thanks{Corresponding author: uct@stjohngrimbly.com} \and
Nicolas Kuske\inst{1,2}\authorcid{0000-0002-1054-6811} \and
Evert A. Boonstra\inst{2} \and
Charel van Hoof\inst{6} \and
Rowan Hodson\inst{3} \and
Benjamin Rosman\inst{4} \and
Ryan Smith\inst{3} \and
Mark Solms\inst{2}\authorcid{0000-0003-4828-3882}\thanks{Co-senior authors.} \and
Jonathan P. Shock\inst{1,2,5}\authorcid{0000-0003-3757-0376}\protect\footnotemark[\value{footnote}]
}
\authorrunning{Grimbly et al.}
\institute{
Dept.\ of Mathematics \& Applied Mathematics, Univ.\ of Cape Town, South Africa \and
Neuroscience Institute, Univ.\ of Cape Town, South Africa \and
Laureate Institute for Brain Research, Tulsa, OK, USA \and
MIND Institute and CSAM, University of the Witwatersrand, South Africa \and
INRS, Montreal, Canada \and
Delft University of Technology, Dept.\ of Cognitive Robotics
}

\maketitle
\AddToShipoutPictureFG*{\put(\LenToUnit{\dimexpr 1in+\oddsidemargin\relax},\LenToUnit{14mm}){\parbox[b]{\textwidth}{\scriptsize\raggedright This version of the contribution has been accepted for publication, after peer review (when applicable) but is not the Version of Record and does not reflect post-acceptance improvements, or any corrections. The Version of Record will be published in the IWAI 2026 proceedings by Springer Nature. Use of this Accepted Version is subject to the publisher's Accepted Manuscript terms of use \url{https://www.springernature.com/gp/open-research/policies/accepted-manuscript-terms}.}}}

\begin{abstract}
An embodied agent must prioritise among competing bodily needs. In active inference, it can respond to an urgent need by increasing the precision of information about that need or by increasing its preference for outcomes that relieve it. We compare these two interventions, separately and together, in a foraging gridworld with three lethal needs, and ask how each changes survival, perception and the body model the agent learns. Precision sharpens interoceptive posteriors and improves survival. Preference reweighting keeps bodily states further from their lethal boundaries and changes the states from which the agent learns, so its learned model covers fewer need levels, with lower quality across all levels but slightly higher quality on the levels it visits.
    An experiment on 72 further layouts does not establish the predicted preference benefit on difficult layouts, but both interventions improve average survival there and their interaction is positive.
Survival alone therefore gives an incomplete account of how a model prioritises needs and learns about its body.

\keywords{active inference \and interoception \and likelihood precision \and
prior preferences \and expected free energy \and Dirichlet learning \and
homeostasis}
\end{abstract}

\section{Introduction}
\label{sec:intro}

A multi-need embodied agent cannot service every drive at once. Hunger, thirst and the need to breathe can each demand a different action, and survival depends on which takes priority. The agent must also infer how urgent these needs are from its interoceptive signals. Active inference allows us to model this relationship between inference over bodily states and homeostatic control~\cite{seth2013interoceptive,barrett2015interoceptive,seth2016active,allen2018body}. But responding to an urgent need can involve different mechanisms. We ask what difference the choice of mechanism makes to behaviour and learning.

We compare two approaches to prioritising a bodily need. An agent that believes itself thirsty could increase the precision of its thirst signal, making information about thirst more reliable at the expense of other needs. Alternatively, it could increase its preference for water, so that policies leading to water become more attractive. The first approach is \emph{likelihood-precision routing}, related to accounts of attention as precision-weighted sensory evidence~\cite{feldman2010attention,parr2017uncertainty,bastos2012canonical} and to context-dependent likelihood precision in discrete-state visual attention~\cite{mirza2019introducing}. The second is \emph{preference reweighting}, drawing on motivational and homeostatic accounts of active inference~\cite{pezzulo2015active,pezzulo2018hierarchical,tschantz2022simulating}.

These are not two implementations of a single mechanism. Both alter policy selection through expected free energy, but they need not change learning in the same way. Our precision intervention changes both the observations emitted by the body and the likelihood used for inference and planning. Preference reweighting changes which outcomes the agent seeks. Although preferences do not enter the likelihood update directly, the actions they favour change the agent's experience and hence what it learns. Comparing the learned body model therefore tells us something that a survival comparison alone cannot.

The choice also matters for biological interpretation. Precision points to gain
control on interoceptive afferents~\cite{quattrocki2014,seth2013interoceptive},
while preference points to changes in regulated
outcomes~\cite{sterling2012allostasis,pezzulo2015active}. That the same behaviour
can be produced by distinct underlying mechanisms is the familiar problem of
degeneracy in biological systems~\cite{edelman2001degeneracy,prinz2004similar}.
A model that reproduces an animal's behaviour does not fix which mechanism the animal uses.

We test each intervention separately and together in the same foraging agent. To our knowledge, these two ways of responding to interoceptive need have not previously been varied together in the same task while measuring their effects on the learned body model. We relate survival to posterior uncertainty, bodily regulation and the range of states represented in that model.  We then test the most uncertain of these findings, the dependence of the preference effect on layout difficulty, on an independently selected set of layouts.  The question is not only whether the agent survives, but how its way of prioritising needs changes what it perceives, does and learns.

\section{Related Work}
\label{sec:related}

Smith et al.~\cite{smith2020failure} dissociate interoceptive sensory precision from a bodily-state prior in a cardiac-perception task. We adopt the same strategy of varying distinct model parameters, here comparing likelihood precision with preferences. The distinction matters because a state prior expresses a belief about the body, whereas a preference specifies observations the agent \emph{wants}.

\begin{sloppypar}
Precision-as-attention accounts describe attention through the precision of sensory evidence or
prediction errors~\cite{feldman2010attention,parr2017uncertainty,bastos2012canonical}.
In a discrete-state visual-search model, context-dependent likelihood precision has been used to model selective attention~\cite{mirza2019introducing}. The motivational and homeostatic literature encodes desirable bodily
or contextual outcomes through prior
preferences~\cite{pezzulo2015active,pezzulo2018hierarchical,tschantz2022simulating}.
Related homeostatic reinforcement learning
defines reward as reduced physiological deficit~\cite{keramati2014}.

Active inference also admits other ways to change behaviour. Hesp et al.~\cite{hesp2021deeply} derive valence from the dynamics of policy precision, rather than the likelihood precision studied here. Other accounts couple preferences and attentional precision~\cite{pezzulo2018hierarchical,parviziwayne2024preferences}. Our comparison makes the two interventions separately controllable within the discrete active-inference framework synthesised by Da Costa et al.~\cite{dacosta2020synthesis}.

Selective precision allocation is studied in \selfcitepred{}, while Torresan et al.~\cite{torresan2025preferences} compare forms of preference without varying likelihood precision. Neacsu et al.~\cite{neacsu2022structure} show that weakening prior preferences can alter likelihood learning. We build on this indirect relationship by comparing how precision and preference, each responding to interoceptive need, change the experience from which the agent learns.
\end{sloppypar}

\begin{figure}[!b]
  \centering
  \includegraphics[width=0.56\textwidth]{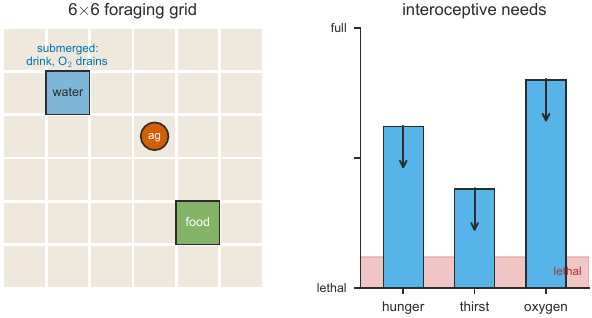}
  \caption{\textbf{The foraging task} (illustrative, with layouts varying): one food and one
  water tile on a $6\times6$ grid. The needs panel (right) tracks three interoceptive
  levels and their lethal floors. Hunger and thirst decline over time. Drinking
  restores thirst but lowers oxygen, which recovers only on land. The best
  priority therefore depends on the current body state.}
  \label{fig:env}
\end{figure}

\section{Methods}
\label{sec:methods}

\subsection{A foraging task with three lethal needs in tension}
\label{sec:env}

We study interoceptive prioritisation in a minimal model of embodied homeostatic
regulation. A \emph{body} has internal variables to maintain, an \emph{environment}
holds resources to replenish them, and an active-inference \emph{agent} senses
the body through interoception and acts to keep it alive. The body couples to the
environment through the agent, so survival is a closed perception--action loop.

The agent forages in \emph{AffectWorld}, a $6\times6$ gridworld with hunger,
thirst and oxygen. Hunger and thirst fall by one level per step. Eating restores
hunger, while drinking restores one third of the thirst range. Water submerges
the agent and lowers oxygen by two levels per step, whereas on land oxygen
recovers by one level per step. Food and water tiles remain fixed and reusable, so the
resource map does not deplete or change during these experiments. The agent dies
when any need crosses its lethal boundary. Drinking can therefore relieve
thirst while endangering oxygen, which places all three needs in tension
(Fig.~\ref{fig:env}).\footnote{The model studied
here has three interoceptive channels ($M=3$). It descends from \selfcitepred{}, which has
four channels. The fourth, a sleep/fatigue channel, and its tiles are removed from
the model rather than disabled.}

\subsection{An active-inference forager that learns its body model online}
\label{sec:agent}

We model the task as a discrete partially observable Markov decision process (POMDP) and use active inference to infer states and select actions~\cite{friston2017active,dacosta2020synthesis}. The POMDP's state factors are grid
position, a context factor held fixed in these experiments, and the true level of
each of the three needs. Position is observed exactly. The current tile's resource
cue is observed with $0.95$ accuracy, and each need produces a noisy ordinal
interoceptive observation. The agent therefore infers only the need levels. Actions combine a move with an optional interaction.
Movement, body dynamics and the fixed resource map are given to the agent. Only the
interoceptive observation likelihood is learned online.

Writing hidden states as $s$, observations as $o$ and policies as $\pi$, the
generative model factorises over a horizon as
\begin{equation}
P(o_{1:T},s_{1:T},\pi) = P(\pi)\,P(s_1)
\prod_{t=1}^{T} P(o_t\mid s_t)
\prod_{t=2}^{T} P(s_t\mid s_{t-1},\pi).
\label{eq:genmodel}
\end{equation}
Here the categorical likelihood is $P(o_t\mid s_t)=\mathrm{Cat}(A)$, transitions are
$P(s_t\mid s_{t-1},\pi)=\mathrm{Cat}(B)$, $C$ contains raw log-preference scores over
observations, and $D$ is the initial-state prior. The corresponding preference
distribution is $P_C=\sigma(C)$. The interoceptive likelihood factorises over the $M=3$ channels,
$A=\{A^{(m)}\}$. The agent infers a posterior $q(s)$ by fixed-point iteration and
scores each policy by its expected free energy,
\begin{equation}
G(\pi) = \underbrace{D_{\mathrm{KL}}\!\big[Q(o\mid\pi)\,\Vert\,P_C(o)\big]}_{\text{risk}}
       + \underbrace{\mathbb{E}_{Q(s\mid\pi)}\big[H[P(o\mid s)]\big]}_{\text{ambiguity}},
\label{eq:efe}
\end{equation}
where $Q(o\mid\pi)=\sum_s A\,Q(s\mid\pi)$ and actions follow $P(\pi)=\sigma(-\gamma\,G(\pi))$.
Up to the normalisation constant in $P_C$, $-G(\pi)$ equals
$\mathbb{E}_{Q(o\mid\pi)}[C(o)] + I_Q(s;o\mid\pi)$, the expected utility plus
the state information gain. Thus $C$ enters expected utility, while $A$ enters predicted
outcomes and the state information gain
$I_Q(s;o\mid\pi)=H[Q(o\mid\pi)]-\mathbb{E}_{Q(s\mid\pi)}H[P(o\mid s)]$.
All four conditions keep this state information gain and omit parameter information gain over the learned Dirichlet counts, so no agent chooses actions in order to improve its body model. What differs between conditions is the experience that their policies provide for learning.

Planning enumerates policies to a horizon of three steps. The agent accumulates Dirichlet counts $\mathbf{a}$
over $A$ and updates them after each observation~\cite{dacosta2020synthesis},
\begin{equation}
  \mathbf{a}^{(m)} \mathrel{+}= \eta\,\big(\mathbf{o}^{(m)} \otimes q(s^{(m)})\big),
  \label{eq:dirichlet}
\end{equation}
where $\mathbf{o}^{(m)}$ is the one-hot observation on channel $m$,
$q(s^{(m)})$ is its state posterior, $\eta$ is the learning rate and $\otimes$
is the outer product. Each observation therefore updates the model of that interoceptive channel, weighted by the inferred body state.  Code, configurations and the supplementary material, which gives the constants, procedures and additional analyses, are available at \nolinkurl{\codeurl}.

\subsection{Likelihood precision and preference reweighting}
\label{sec:mechanisms}

\begin{figure}[t]
  \centering
  \includegraphics[width=0.84\textwidth]{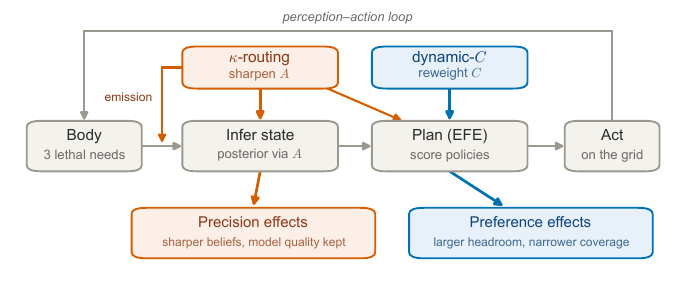}
  \caption{\textbf{The perception--action loop and the two implemented mechanisms.} Precision
  ($\kappa$) sharpens the likelihood $A$, which feeds inference and planning and governs
  the body's emission. Preference reweighting changes $C$, which enters only the
  planner's pragmatic term. The lower boxes show principal internal effects on the original layouts, not effects exclusive to either intervention.}
  \label{fig:loop}
\end{figure}

Both interventions respond to the agent's posterior estimates of need. Precision routing gives more reliable information about the need with the largest expected deficit. Preference reweighting instead uses the estimated hunger and thirst deficits to set preferences for food and water (Fig.~\ref{fig:loop}).

\paragraph{Likelihood precision (\kapparoute{}).} A scalar precision $\kappa$ on
the diagonal of the interoceptive likelihood $A^{(m)}$ controls how reliably a
true need level is reported. The same allocation changes the body's emitted observations and the likelihood used for inference and planning. At higher $\kappa$ the observation model places more
mass on the correct ordinal level, and at lower $\kappa$ it spreads that mass over
the alternatives. Thus, when thirst is selected, the agent receives more reliable
evidence about thirst and less reliable evidence about hunger and oxygen. Here
$\kappa$ is the diagonal probability mass of the likelihood, not an
inverse-variance precision. The total precision across channels is fixed at
$K = M\cdot\bar\kappa$, so prioritising one
channel necessarily de-prioritises the others. Because
$\kappa$ scales $A$, it affects state inference, policy evaluation and the
Dirichlet update in Eq.~\eqref{eq:dirichlet}. This is the precision-routing
mechanism of \selfcitepred{}.

\paragraph{Preference reweighting (\dynC{}).} This intervention
rewrites the raw resource preference scores $C$ from the posterior deficits and
does not directly alter $A$. Greater expected thirst makes water more desirable, and greater
expected hunger does the same for food. If either deficit crosses the critical
threshold, the alternative resource and empty tiles receive a large penalty. Because $C$ appears only in the pragmatic part of the policy
score, this intervention changes action selection without entering the likelihood update
in Eq.~\eqref{eq:dirichlet}.

Oxygen exposes an asymmetry between the interventions. It recovers passively on land and has no
navigable resource cue, so preference reweighting does not target it directly.
It can protect oxygen only by favouring policies that leave or avoid water.
Precision routing can instead assign precision to the oxygen channel itself.
The survival comparison therefore concerns the two interventions as implemented in a
mixture of actionable and passive needs. It does not isolate their performance
under symmetric access to every channel.

\label{sec:design}

We compare four conditions using the same agent architecture. The \baseuniform{} keeps $\kappa$ uniform and $C$ static. The \kapparoute{} condition changes precision alone, \dynC{} changes preferences alone, and the combined condition applies both interventions.

\subsection{What changes in behaviour and learning?}
\label{sec:fingerprint}
\label{sec:metrics}

Survival tells us whether prioritisation helps the agent stay alive. To understand how it does so, we also examine the agent's uncertainty, bodily states and learned likelihood. In particular, preference can alter learning by changing the states the agent visits, even though it does not appear explicitly in Eq.~\eqref{eq:dirichlet}. We distinguish the quality of the learned model across all need levels from its quality on visited levels, and measure the fraction of levels visited. This tests whether a lower overall score accompanies poorer learning on visited levels or narrower experience.

The \emph{survival
fraction} is the proportion of the 120-step episode completed. Comparisons pair
the same seed and layout across conditions.

On the perceptual side, \emph{posterior entropy}
$\bar H = \big\langle -\sum_s q(s^{(m)})\ln q(s^{(m)})\big\rangle$ measures
uncertainty about the need levels, averaged over channels and surviving steps in the second half of the episodes. Lower entropy means a sharper posterior.
\emph{Learned-model quality} is the probability that the learned likelihood
assigns to the observation favoured by the true model, averaged over latent levels,
$p_\star^{(m)} = \big\langle \hat A^{(m)}\!\big[\arg\max_{o} A^{(m)}_{o,s},\,s\big]\big\rangle_s$.
We report this both across all levels and across
levels the agent visits.

On the behavioural side, \emph{visited-state entropy} $H[p_{\mathrm{visit}}]$
measures the breadth of body states occupied. The \emph{minimum headroom}
$b = \big\langle \min_m s^{(m)}_t/L_m\big\rangle_t$ is the mean distance of the
most endangered need from its lethal boundary, expressed as a fraction of that
channel's range. It is calculated from true body levels, so sharper perception
cannot change the measurement itself.

\subsection{Experiments and analysis}
\label{sec:experiments}

The original experiment compares the four conditions on eleven distinct layouts with eight paired seeds. Each run comprises 20 learning episodes of 120 steps, and the learned-model comparison uses the same layouts and seeds over a longer run. These layouts were selected in an earlier task to distinguish controllers that respond differently to urgency, so we measured their difficulty in the present task from baseline survival rather than inheriting labels from that selection.

Intervals resample layouts while preserving the paired comparisons~\cite{efron1993introduction}. For the three original survival contrasts, we use exact two-sided sign-flip tests with Holm correction~\cite{holm1979simple}. The internal-measurement intervals are not adjusted for multiple comparisons. With only eleven independent layouts, the original intervals are wide.

We also test whether the preference benefit depends on difficulty, as suggested by a post-hoc split of the original layouts. This experiment uses 72 additional layouts selected without inspecting intervention outcomes. Separate baseline runs assign them to three difficulty groups of 24, and evaluation uses eight new paired seeds. The primary test asks whether preference improves survival on the difficult layouts, averaging its effect with and without precision routing. The continuous relationship with difficulty and the effects across all 72 layouts are secondary analyses. The design and analysis were fixed before evaluation outcomes were inspected.

\section{Results}
\label{sec:results}

\begin{figure}[tbp]
  \centering
  \includegraphics[width=0.86\textwidth]{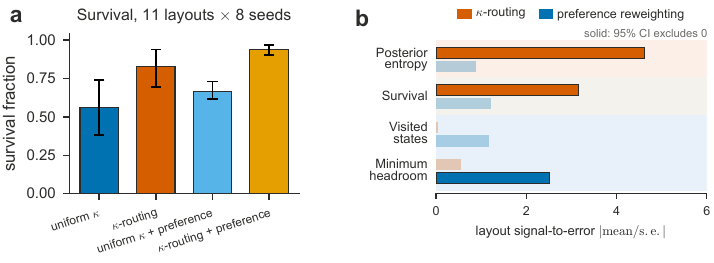}
  \caption{\textbf{Survival and the internal dissociation.} \textbf{(a)} Condition means
  and layout-bootstrap $95\%$ intervals across eleven layouts and eight seeds.
  \textbf{(b)} Layout-level signal-to-error ($|\mathrm{mean}/\mathrm{s.e.}|$).
  Bar length is descriptive. Solid bars have $95\%$ intervals excluding zero
  (Tables~\ref{tab:stats}, \ref{tab:internal}).}
  \label{fig:results}
\end{figure}

\subsection{How does each intervention change survival?}
\label{sec:res-survival}
\begin{sloppypar}
With the \baseuniform{}, the agent survives $0.56$ of the episode, or about 67 of 120
steps. \kapparoute{} raises survival to $0.83$, and the combined condition reaches
$0.94$. Preference reweighting alone reaches $0.67$ (Fig.~\ref{fig:results}a and
Table~\ref{tab:stats}). The direction of the precision result agrees with \selfcitepred{}. Precision and the combined intervention improve survival after correction for multiple comparisons. The preference-alone estimate is less certain. Its interval includes both a decrease and an increase in survival, and an equivalence test is also inconclusive.
\end{sloppypar}

\begin{table}[t]
  \centering
  \caption{Survival contrasts on the eleven original layouts with eight paired seeds. Intervals are paired layout
  bootstraps. The $p$ values use exact layout-level sign flips with Holm
  adjustment across the three contrasts.}
  \label{tab:stats}
  \begin{tabular}{lccc}
    \toprule
    Contrast & $\Delta$survival & 95\% CI & Holm $p$ \\
    \midrule
    \kapparoute{} vs \baseuniform{}             & $+0.27$ & $[0.12,\ 0.44]$ & $0.0176$ \\
    \dynC{} vs \baseuniform{}                    & $+0.11$ & $[-0.06,\ 0.27]$ & $0.248$ \\
    \kapparoute{}+\dynC{} vs \baseuniform{}      & $+0.38$ & $[0.18,\ 0.56]$ & $0.0176$ \\
    \bottomrule
  \end{tabular}
\end{table}

In a post-hoc split of the original layouts, preference alone improved survival on the five where the baseline struggled and reduced it on the other six. This suggested that its benefit might depend on difficulty. We tested that possibility on the additional layouts, this time averaging the preference effect across both precision conditions rather than testing preference alone.
\begin{sloppypar}
\paragraph{Does the difficulty pattern generalise?}
The predicted preference benefit on the layouts calibrated as difficult was not established (Table~\ref{tab:confirm-main}). The primary estimate is close to zero, with an interval spanning both signs.

The continuous analysis also does not establish the predicted decrease in benefit with increasing baseline survival.
The difficulty ranking is limited, however, by 51 tied baseline-calibration values.

Across all 72 layouts, the planned secondary analyses show positive average survival effects for both interventions and a positive interaction (Table~\ref{tab:confirm-main}). These results support complementarity on the survival-fraction scale.

On these additional layouts, the internal measurements show the same dominant pattern as on the original layouts. \kapparoute{} lowers posterior entropy, and \dynC{} raises headroom and lowers conflict, although each also changes measurements associated with the other intervention.
\end{sloppypar}

\begin{table}[t]
  \centering
  \small
  \caption{Survival effects in the additional experiment. The primary test uses the 24 layouts calibrated as difficult. Secondary effects average over all 72 layouts. Intervals resample layouts. Secondary $p$ values are Holm-adjusted jointly with the continuous difficulty test.}
  \label{tab:confirm-main}
  \begin{tabular}{@{}p{0.47\textwidth}ccc@{}}
    \toprule
    Comparison & Effect & 95\% CI & $p$ \\
    \midrule
    Preference, difficult layouts (primary) & $\ConfirmPrimaryMean$ & $[\ConfirmPrimaryCILow,\,\ConfirmPrimaryCIHigh]$ & $\ConfirmPrimaryP$ \\
    \midrule
    Precision, all layouts (secondary) & $\ConfirmMarginalKappaMean$ & $[\ConfirmMarginalKappaCILow,\,\ConfirmMarginalKappaCIHigh]$ & $\ConfirmMarginalKappaHolmP$ \\
    Preference, all layouts (secondary) & $\ConfirmMarginalCMean$ & $[\ConfirmMarginalCCILow,\,\ConfirmMarginalCCIHigh]$ & $\ConfirmMarginalCHolmP$ \\
    Interaction, all layouts (secondary) & $\ConfirmInteractionMean$ & $[\ConfirmInteractionCILow,\,\ConfirmInteractionCIHigh]$ & $\ConfirmInteractionHolmP$ \\
    \bottomrule
  \end{tabular}
\end{table}

\subsection{Precision sharpens perception, preference changes the learned model}
\label{sec:res-acceleration}
On the original layouts, precision sharpens the interoceptive posterior
(Table~\ref{tab:internal}, Fig.~\ref{fig:results}b).
The preference contrast is smaller and its interval includes zero, so these data do not resolve a change in posterior entropy.

The learned-model comparison reveals why it matters which states an agent visits. Preference reweighting lowers model quality when averaged over all need levels, yet slightly improves quality on the levels actually visited. It also visits fewer levels. Those it rarely visits remain close to the vague prior and lower the all-level average. In other words, the lower score accompanies narrower experience rather than poorer performance on the states encountered. Precision shows no comparable drop in all-level quality in the same learning runs.

\begin{table}[!t]
  \centering
  \small
  \caption{Internal contrasts against baseline, shown as mean [layout-bootstrap $95\%$ CI].
  Entropies use $10^{-3}$ nats (posterior) and nats (visited state). Headroom is a fraction.}
  \label{tab:internal}
  \begin{tabular}{lccc}
    \toprule
    Metric (vs.\ baseline) & \kapparoute{} & \dynC{} & \kapparoute{}+\dynC{} \\
    \midrule
    Posterior entropy ($10^{-3}$) & $-1.37\,[\text{-}1.95,\text{-}0.84]$ & $-0.26\,[\text{-}0.83,0.28]$ & $-1.18\,[\text{-}1.77,\text{-}0.58]$ \\
    Visited-state entropy         & $+0.00\,[\text{-}0.09,0.07]$ & $-0.05\,[\text{-}0.14,0.02]$ & $-0.15\,[\text{-}0.24,\text{-}0.08]$ \\
    Minimum headroom              & $+0.01\,[\text{-}0.02,0.04]$ & $+0.05\,[0.01,0.09]$ & $+0.11\,[0.07,0.15]$ \\
    \bottomrule
  \end{tabular}
\end{table}

\subsection{Preference keeps needs further from their lethal boundaries}
\label{sec:res-regulation}
Preference increases minimum headroom on the original layouts by $+0.05$, and the combined condition by
$+0.11$ (Table~\ref{tab:internal}). The precision contrast is $+0.01$, with an interval that includes zero. Both interventions
also reduce the time spent with two needs simultaneously near their lethal boundaries, from about $2\%$
of baseline steps to less than $0.1\%$, a comparison we report without an interval. They reach this low-conflict state in
different ways. Precision sharpens the evidence about need, while preference keeps the most endangered need further from its boundary.

\subsection{Robustness on the original layouts}
\label{sec:res-robustness}

Replacing posterior deficits with true body levels leaves preference survival
unchanged on every original layout. Sweeping the critical penalty over two
orders of magnitude and removing its threshold also preserves the easy-layout
cost, as do held-out seeds. The original pattern is therefore
not explained by posterior error or one penalty setting, although it did not generalise to the additional layouts.

A belief-only $\kappa$ variant keeps emitted observation precision uniform while routing precision during inference and planning. On six of the original layouts it improves survival by $+0.33$. The full-intervention contrast of $+0.27$ in Table~\ref{tab:stats} uses the original eleven layouts, so these estimates are not a paired decomposition of the emission effect. The subset result suggests that sharper emission is not required for a benefit, while the primary intervention remains composite.

\section{Discussion}
\label{sec:discussion}

The choice of prioritisation mechanism changes more than survival. Precision alters the evidence the agent receives and how it uses that evidence in inference and planning, while preference changes which outcomes the agent seeks. Both also change where the agent goes, and so what it later observes, believes and learns.  On the additional layouts preference raised posterior entropy and precision slightly lowered headroom, so neither intervention is confined to perception or to action.

Why does preference change the learned likelihood when it never enters the learning rule? The answer is the agent's experience. An agent that commits to relieving one need visits a narrower range of bodily states, and the levels it rarely visits stay close to the prior. It learns those states no less accurately, only less often. This is an interoceptive instance of the indirect coupling that Neacsu et al.~\cite{neacsu2022structure} describe, and it is why an average over the whole likelihood says little about what the agent has learned from the states it actually encountered.

On the original layouts, preference helped where the baseline struggled and cost survival where the baseline already coped. Reweighting appears to overcommit the agent to one need, which is useful when the agent is failing and harmful when it is not.
    The additional layouts did not reproduce this pattern. Preference did not help on the layouts calibrated as difficult, no continuous relation with baseline survival was detected, and the unadjusted tertile contrasts run in the opposite direction, with the largest benefit on the easiest layouts. We therefore keep the overcommitment account for the original layouts only.
  What the additional layouts do show is that both interventions improve survival on average, and that the two together improve it by more than the sum of their separate effects. That is evidence of complementarity in survival, though not of how the two mechanisms combine.

This matters for how active inference is used to model biological systems. Fitting behaviour does not tell us whether prioritisation works through interoceptive gain or through preferred outcomes, because both can produce the same survival. The internal measurements suggest what would. Uncertainty or gain, distance from physiological limits and the range of states experienced can all be measured alongside behaviour. We compare two implemented mechanisms rather than testing formal parameter identifiability, and the same ambiguity applies to homeostatic reinforcement learners~\cite{keramati2014}.

The original experiment uses eleven selected layouts and its difficulty split is post hoc.
  The additional experiment separates calibration from evaluation but is still one task.
Preference also has no direct target for oxygen, which recovers passively. A symmetric comparison would need actionable oxygen or preferences over body outcomes, and the joint hunger--thirst gate of the preference rule was never triggered in these runs.

Precision could also act on transition beliefs or on policy selection~\cite{limanowski2024precision}. Varying emitted and believed noise separately would show which part of the present precision intervention carries its effect. More channels, richer environments and longer planning horizons would show how far the results extend beyond this gridworld and its three-step planner.

\section{Conclusion}
\label{sec:conclusion}

Likelihood precision and preference reweighting provide distinct mechanisms for interoceptive prioritisation, with different dominant effects on posterior uncertainty, bodily regulation and state coverage.
    The predicted preference benefit on independently calibrated difficult layouts was not established.
  Across all 72 layouts, the secondary analyses show positive average survival
  effects for both interventions and a positive interaction.
Their internal observables reveal effects on later inference and learning that survival alone does not resolve. Comparing these observables is therefore important when using behavioural success to evaluate a model of prioritisation.

\section*{Declarations}
\begin{credits}

\subsubsection{Acknowledgements}
Funded by the Oppenheimer Memorial Trust (UCT Neuroscience Institute,
475201 NSI1006) and Conscium, Ltd (UCT Psychology, PSY526 428430). EAB: NWO
grant 019.223SG.002. RS: Laureate Institute for Brain Research.

\subsubsection{Use of AI tools}
AI-based tools assisted with writing, code development and research planning.
The corresponding author verified all results, claims and code.

\begin{sloppypar}
\subsubsection{Data and code availability}
The project repository at \nolinkurl{\codeurl} contains the scoped data,
code and exact environment needed to reproduce the reported experiments and analyses, together with the supplementary material.
\end{sloppypar}

\subsubsection{Disclosure of Interests}
The authors have no competing interests to declare.
\end{credits}

\clearpage
\bibliographystyle{splncs04}
\bibliography{paper2_refs}

\clearpage


\section*{Supplementary material}

# Two Routes to Interoceptive Prioritisation: Dissociating Likelihood Precision and Preference Reweighting in Active Inference

## Supplementary Material

St John Grimbly[1,2,⋆], Nicolas Kuske[1,2], Evert A. Boonstra[2], Charel van Hoof[6], Rowan Hodson[3], Benjamin Rosman[4], Ryan Smith[3], Mark Solms[2⋆⋆], and Jonathan P. Shock[1,2,5⋆⋆]

[1] Dept. of Mathematics & Applied Mathematics, Univ. of Cape Town, South Africa
[2] Neuroscience Institute, Univ. of Cape Town, South Africa
[3] Laureate Institute for Brain Research, Tulsa, OK, USA
[4] MIND Institute and CSAM, University of the Witwatersrand, South Africa
[5] INRS, Montreal, Canada
[6] Delft University of Technology, Dept. of Cognitive Robotics

## A Model and constants

This section states the parts of the model that the main paper summarises.

*Interoceptive likelihood.* Each channel $m$ has a latent level $s \in \{0, \ldots, L_m\}$, with $L_m = 10$ for hunger and thirst and $L_m = 6$ for oxygen. The agent observes $n = 6$ ordinal need levels $\beta_m(s) = \mathrm{round}\big((n-1)(L_m - s)/L_m\big)$, and the likelihood places mass $\kappa_m$ on the reported level,

$$A^{(m)}_{o,s} = \begin{cases} \kappa_m & o = \beta_m(s), \\ (1-\kappa_m)/(n-1) & o \neq \beta_m(s). \end{cases} \tag{S1}$$

Under the budget $K = M\bar{\kappa} = 1.95$ the attended channel receives $\kappa = 0.9$ and the other two 0.525 each. The uniform allocation gives 0.65 to every channel.

*Precision selector.* The attended channel is the one the agent believes closest to its lethal boundary, with ties broken by the lowest index,

$$m^\star = \arg\max_m \; \mathbb{E}_{Q(s^{(m)})}\big[(L_m - s)/L_m\big]. \tag{S2}$$

*Preference reweighting.* At each step dynamic-$C$ sets the preferences for food and water from the believed deficits $t_m = \max(0,\, \bar{L} - \mathbb{E}_{Q(s^{(m)})}[s])$, with a fixed

⋆ Corresponding author: uct@stjohngrimbly.com
⋆⋆ Co-senior authors.

Table S1: Constants, identical across the four conditions.

| Quantity | Symbol | Value |
|---|---|---|
| Interoceptive channels | $M$ | 3 (hunger, thirst, oxygen) |
| Ordinal levels per channel | $n$ | 6 |
| Base precision | $\bar{\kappa}$ | 0.65 |
| Attended precision cap | $\kappa_{\mathrm{att}}$ | 0.90 |
| Precision budget | $K = M\bar{\kappa}$ | 1.95 |
| Background floor | $\kappa_{\min}$ | 0.05 |
| Deficit reference | $\bar{L}$ | 15 |
| Critical threshold | $\tau$ | 14 |
| Preference critical penalty | $P_{\mathrm{crit}}$ | −500 |
| Policy precision | $\gamma$ | 1 (`pymdp` default) |
| Dirichlet learning rate | $\eta$ | 1.0 |
| Dirichlet concentration | $\alpha_0$ | 0.1 |
| Planning horizon | $H$ | 3 (full enumeration) |
| Grid | — | $6 \times 6$, one food and one water tile |
| Original layouts | — | 12 IDs (11 layouts), 8 seeds; $20 \times 120$ |
| Learned-model run | — | 12 IDs (11 layouts), 8 seeds; $30 \times 150$ |
| Additional layouts | — | 72 layouts; calibration 4 seeds, evaluation 8 seeds; $20 \times 120$ |
| Resamples | — | $10^4$ bootstraps (original), $2 \times 10^4$ bootstraps and $10^5$ permutations (additional) |

reference $\bar{L} = 15$ for both hunger and thirst. Because $\bar{L}$ exceeds their ceiling of 10, each deficit lies between 5 and 15:

$$C^{\mathrm{res}}[\text{food}] = \begin{cases} P_{\mathrm{crit}} & t_{\mathrm{thirst}} > \tau, \\ t_{\mathrm{hunger}} & \text{otherwise}, \end{cases} \qquad C^{\mathrm{res}}[\text{water}] = \begin{cases} P_{\mathrm{crit}} & t_{\mathrm{hunger}} > \tau, \\ t_{\mathrm{thirst}} & \text{otherwise}. \end{cases} \tag{S3}$$

A deficit exceeds $\tau = 14$ only when the expected level is below 1. The other outcome slots receive $P_{\mathrm{crit}}$ when either deficit is critical and an empty bias of −1 otherwise, and poison keeps −4. Oxygen has no navigable resource and is not reweighted. $C$ does not enter the likelihood or its Dirichlet update.

*Original layouts.* The food and water positions were selected in an earlier task without lethal oxygen, to separate an urgency-following controller from one that ignores urgency. The IDs `layout0001` and `layout0068` share coordinates and are averaged before inference, giving 88 paired layout–seed records over eleven layouts. In a post-hoc median split, preference alone improves survival by +0.36 on the five layouts with lower baseline survival and reduces it by −0.11 on the other six. Its overall survival contrast has a 90% interval of $[-0.03, +0.24]$, which does not establish equivalence within ±0.06.

## B Additional layouts

The 72 layouts are the first eligible placements from a fixed seed stream that are distinct under rotation and reflection of the grid and from the original layouts. They were chosen without reference to any agent outcome. Baseline-only runs on seeds 200–203 ranked them into 24 hard, 24 middle and 24 easy layouts. Fifty-one layouts share the modal calibration survival of 0.083, and travel burden, the largest start-to-resource distance and a fixed geometry signature break those ties. The four conditions were then run on seeds 300–307.

For layout $g$, let $\bar{Y}_g(k,c)$ be survival averaged over the eight seeds, where $k$ and $c$ indicate precision routing and preference reweighting. The marginal preference effect is

$$d_g^C = \frac{1}{2}\left\{[\bar{Y}_g(0,1) - \bar{Y}_g(0,0)] + [\bar{Y}_g(1,1) - \bar{Y}_g(1,0)]\right\}, \qquad \text{(S4)}$$

and the primary test is its mean over the 24 hard layouts. By tertile, the marginal preference effect is $-0.008$ on hard, $+0.059$ $[+0.003, +0.122]$ on middle and $+0.190$ $[+0.128, +0.257]$ on easy layouts (unadjusted; Fig. S1). Its slope on calibration survival is $+0.025$ $[-0.102, +0.266]$ (permutation $p = 0.7595$). Using seed medians gives a primary estimate of $-0.011$, and leaving out each hard layout in turn gives means from $-0.018$ to $-0.006$.

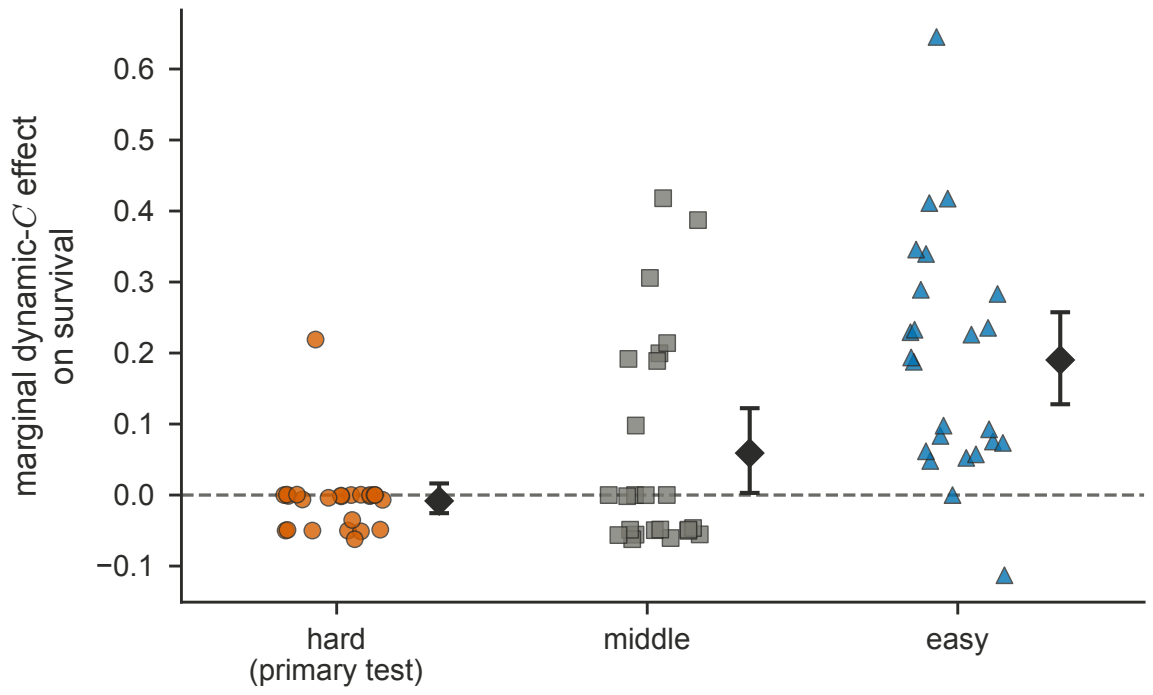


Fig. S1: **Marginal dynamic-$C$ effect on survival for each of the 72 additional layouts.** Points are layouts, averaged over eight seeds and grouped by calibrated tier (24 each). Diamonds are tier means with 95% layout-bootstrap intervals. The hard tier is the primary test.

Among the internal measures, marginal $\kappa$-routing lowers posterior entropy, broadens coverage and slightly lowers headroom. Marginal dynamic-$C$ raises headroom, lowers conflict, narrows coverage and raises posterior entropy. Each of these effects has a 95% interval that excludes zero, while the $\kappa$-routing effect on conflict does not.

## C State coverage and the learned-model measure

Preference reweighting does not enter the Dirichlet update, but it changes which levels the agent visits and so which columns of the likelihood are updated. With preference reweighting the agent visits fewer levels, and it learns the levels it does visit at least as well (Table S2). These values come from a separate learning run of 30 episodes of 150 steps, with the uniform baseline and dynamic-$C$ only, in which hunger and thirst ranged up to 15 rather than 10. The main text's comparison with precision uses an earlier three-layout run of the same runner with all four conditions.

Table S2: Preference effects relative to the uniform baseline over 96 paired layout–seed records on the 12 layout IDs, two of which share coordinates. Quality is averaged over all levels of the three channels. Intervals resample layout IDs.

| Measurement | Difference | 95% CI |
|---|---|---|
| Model quality, all levels | −0.030 | [−0.040, −0.021] |
| Model quality, visited levels | +0.008 | [+0.002, +0.015] |
| Fraction of levels visited | −0.091 | [−0.123, −0.068] |


\end{document}